\documentclass{article}
\usepackage{spconf,amsmath,graphicx}

\usepackage{subfigure}

\usepackage[keeplastbox]{flushend}
\usepackage{epsfig}

\title{Dense Non-rigid Structure-from-Motion Made Easy -- A Spatial-Temporal Smoothness based Solution}
\name{Yuchao Dai$^{1,2}$,  \quad Huizhong Deng$^{1}$, \quad Mingyi He$^{2}$
\thanks{This work is supported in  part by the Australian Research Council Grant (DE140100180) and National Natural Science Foundation of China (61420106007, 61671387). Yuchao Dai (daiyuchao@gmail.com) is the corresponding author.}
}
\address{$^{1}$Research School of Engineering, Australian National University, Australia \\ $^{2}$ School of Electronics and Information, Northwestern Polytechnical University, China}
\newcommand{\rank}{\mbox{\rm rank}}

\newcommand{\SKIP}[1]{} % Used to skip stuff we do not want type-set

\newcommand{\mbegin} {\left [ \begin{array}}

\newcommand{\mend}   {\end{array} \right ]}

\newcommand{\detbegin} {\left | \begin{array}}
\newcommand{\detend}   {\end{array} \right |}

\newcommand{\vbegin} {\left ( \begin{array}{c}}

\newcommand{\vend} {\end{array}\right )}

\def\squareforqed{\hbox{\rlap{$\sqcap$}$\sqcup$}}

\def\qed{\ifmmode\squareforqed\else{\unskip\nobreak\hfil
	\penalty50\hskip1em\null\nobreak\hfil\squareforqed
	\parfillskip=0pt\finalhyphendemerits=0\endgraf}\fi}

\newcommand{\showeqnlabel}{
	\hbox to 0pt{\quad\quad\relax\fbox{\scriptsize\rm\eqnlblx}%
	\gdef\eqnlblx{xxxx}}} \newcommand{\eqnlblx}{}

\def\@eqnnum{\rm (\theequation)\showeqnlabel}

\newcommand{\nofig}[1]{\centerline{\bf Figure here}}

\def\mat#1{\mathchoice{\mbox{\bf$\displaystyle\tt#1$}}
	{\mbox{\bf$\textstyle\tt#1$}}
	{\mbox{\bf$\scriptstyle\tt#1$}}
	{\mbox{\bf$\scriptscriptstyle\tt#1$}}}
\def\m#1{\protect\mat #1}

\newcommand\eg{\emph{e.g.}} 
\newcommand\ie{\emph{i.e.}} 
 
\newcommand\etal{\emph{et al.}} 
\newcommand\etc{\emph{etc.}}

\begin{document}
%\ninept
%
\maketitle
\begin{abstract}
This paper proposes a simple spatial-temporal smoothness based method for solving dense non-rigid structure-from-motion (NRSfM). First, we revisit the temporal smoothness and demonstrate that it can be extended to dense case directly. Second, we propose to exploit the spatial smoothness by resorting to the Laplacian of the 3D non-rigid shape. Third, to handle real world noise and outliers in measurements, we robustify the data term by using the $L_1$ norm. In this way, our method could robustly exploit both spatial and temporal smoothness effectively and make dense non-rigid reconstruction easy. Our method is very easy to implement, which involves solving a series of least squares problems. Experimental results on both synthetic and real image dense NRSfM tasks show that the proposed method outperforms state-of-the-art dense non-rigid reconstruction methods.
\end{abstract}
\section{Introduction}
Non-rigid structure-from-motion (NRSfM) aims at simultaneously recovering the camera motion and non-rigid structure from 2D images by using a monocular camera, which is central to many computer vision applications (3D reconstruction, motion capture, human-computer interaction  etc) and has received considerable attention in recent years. A great number of methods have been established, and most of the existing methods can be roughly categorized as sparse methods and dense methods \cite{Bregler:CVPR-2000} \cite{Xiao-Chai-Kanade:ECCV-2004}\cite{Torresani-Hertzmann-Bregler:PAMI-2008} \cite{metric-projection:CVPR-2009} \cite{Dai-Li-He:CVPR-2012}\cite{Procrustean-Normal-Distribution:CVPR-2013} \cite{Simon2014}.

%The research of non-rigid structure from motion (NRSfM) began with Bregler \etal's seminal work\cite{Bregler:CVPR-2000} on extending structure-from-motion from rigid factorization to non-rigid factorization in a sparse scenario. However, it is soon proved that NRSfM has inherent ambiguities \cite{Xiao-Chai-Kanade:ECCV-2004} in shape basis and shape coefficients. Multiple priors are proposed to address this under-constrained problem: shape space constraints that limit the shape ambiguity\cite{Torresani-Hertzmann-Bregler:PAMI-2008} \cite{metric-projection:CVPR-2009} \cite{Dai-Li-He:CVPR-2012}\cite{Procrustean-Normal-Distribution:CVPR-2013}, trajectory space approaches that focus on temporal smoothness of point trajectories\cite{Akhter-Sheikh-Khan-Kanade:Trajectory-Space-2010}\cite{Park-Trajectory:ECCV-2010}, and a combination of shape and trajectory constraints\cite{Complementary-rank-3:CVPR-2011}\cite{Simon2014}.

NRSfM is in essential under-determined (estimating a 3D point from a single 2D measurement), therefore, extra regularization is needed to constrain the problem. For sparse NRSfM, various priors/constraints have been enforced, such as shape basis \cite{Bregler:CVPR-2000}, trajectory basis \cite{Akhter-Sheikh-Khan-Kanade:Trajectory-Space-2010}, shape-trajectory basis \cite{Complementary-rank-3:CVPR-2011}\cite{Simon2014} and smoothness \cite{Dai-Li-He:IJCV-2013}. In sparse reconstruction, the feature points are geometrically apart from each other, thus spatial regularization cannot be enforced. By contrast, dense NRSfM aims at achieving 3D non-rigid reconstruction for each pixel in a video sequence, where spatial constraint has been widely exploited to regularize the problem \cite{Dense-NRSFM:3DIMPVT-2012}\cite{Video-Registration:IJCV-2013}\cite{Russell2014}. Garg \etal \cite{Dense-NRSFM:CVPR-2013} proposed to enforce both the total variation constraint and the nuclear norm induced low-rank constraint on the 3D non-rigid shape. This results in a complex convex optimization and GPU is needed to speed up the implementation. Furthermore, they only implemented the method on complete and noise-free datasets, thus its robustness remains questionable. In Russell \etal's work\cite{Russell2014}, segmentation is performed on both object-level and part-level, then piece-wise reconstruction is applied by assuming locally rigid pieces. In \cite{Grouping-NIPS2014}, motion segmentation is paired with rank constrained 2D track completion to deal with occlusions, then nuclear norm minimization is used to recover the 3D shape. Yu \etal \cite{Yu_2015_ICCV} proposed to utilize the temporal smoothness in both camera motion and 3D deformation, where a template 3D shape is available. Ranftl \etal\cite{Ranftl_2016_CVPR} investigated the relative scale in dynamic scene. All these constraints are based on motion and semantic segmentation, thus computational complex. 

\begin{figure*}[!htb]
       \subfigure[Input W]{       \psfig{file=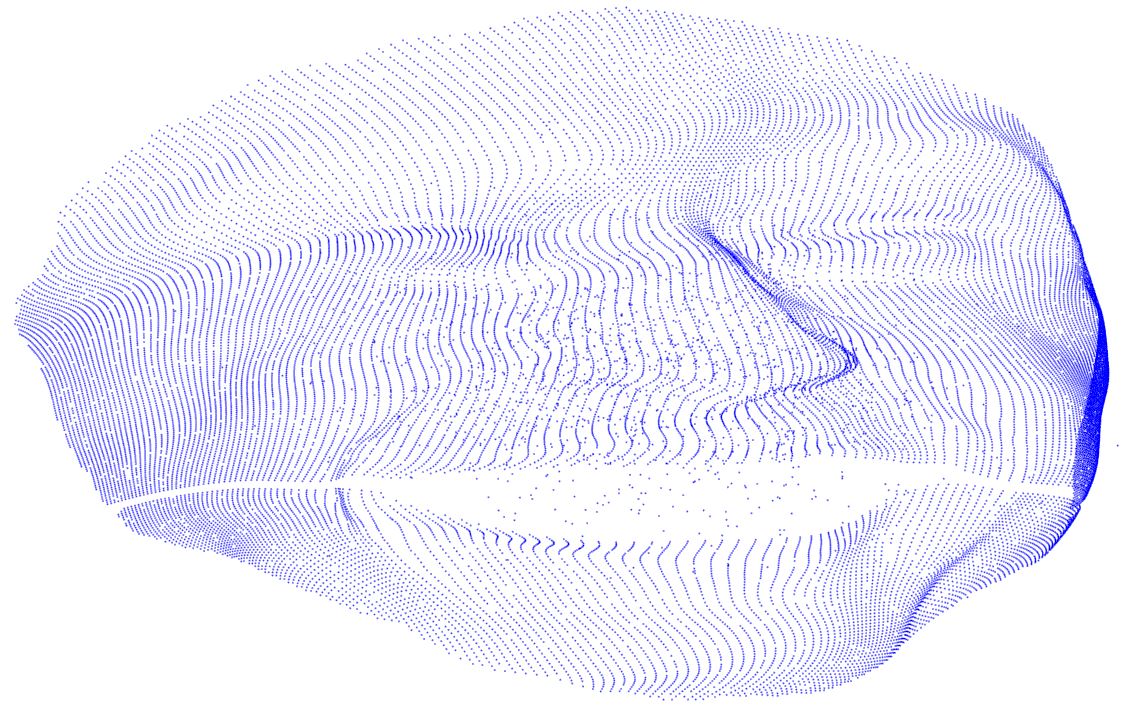,width=0.18\textwidth,
       height=0.11\textheight}
	   }
       \subfigure[Pseudo-inverse (0.5084)]{       \psfig{file=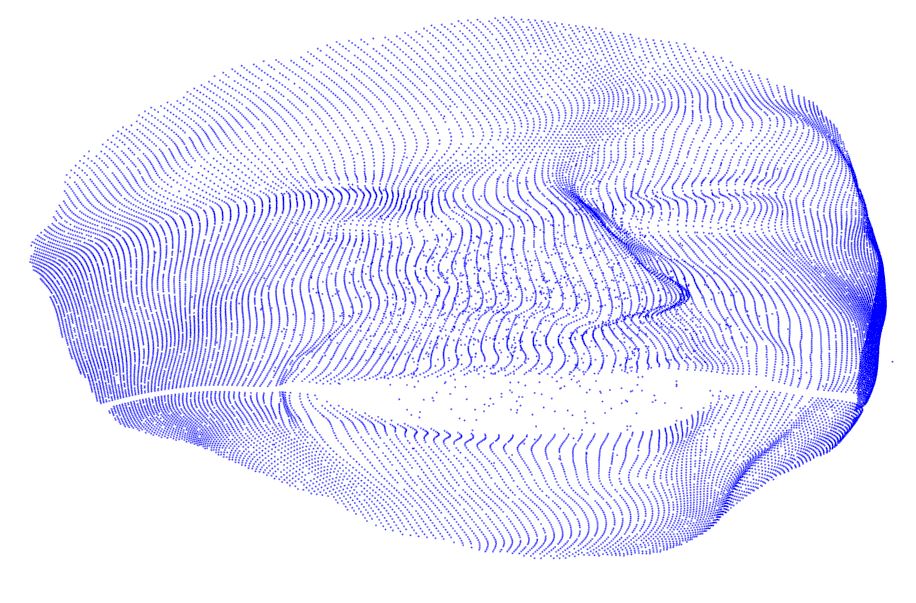,width=0.18\textwidth,
       height=0.11\textheight}
	   }
       \subfigure[Temporal (0.1835)]{       \psfig{file=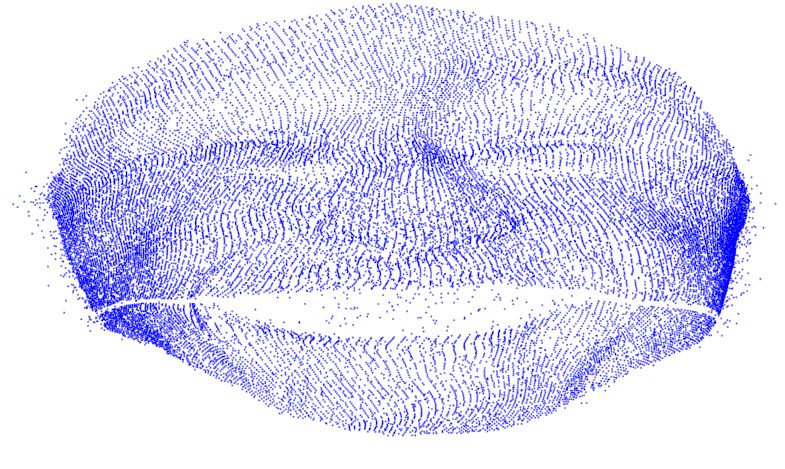,width=0.18\textwidth,
       height=0.11\textheight}
	   }
       \subfigure[Spatial-temporal (0.0757)]{       \psfig{file=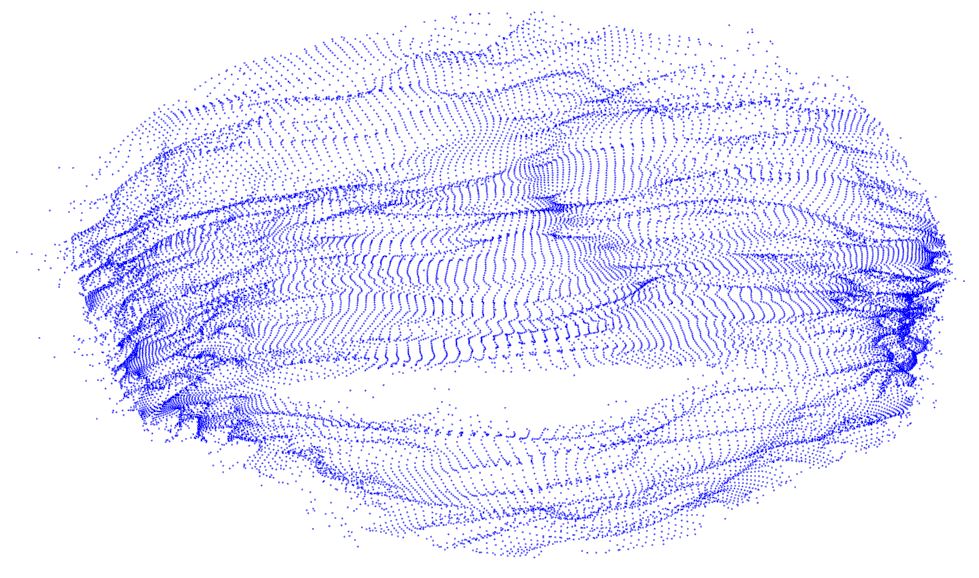,width=0.18\textwidth,
       height=0.11\textheight}
	   }
       \subfigure[Final (0.0700)]{
       \psfig{file=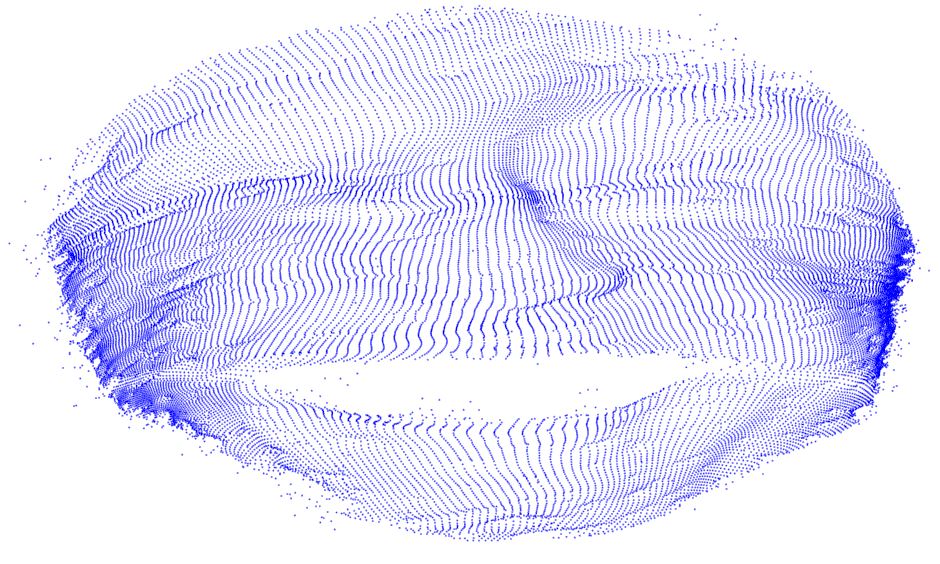,width=0.18\textwidth,
       height=0.11\textheight}
	   }
       \caption{\small A nutshell of our robust temporal-spatial smoothness based dense non-rigid reconstruction, where the experiments are conducted on the Synthetic Face Sequence 2 \cite{Dense-NRSFM:CVPR-2013} with outliers. The brackets include the RMS 3D error of the full 3D reconstruction under each scenario. By enforcing the temporal smoothness, spatial smoothness and applying our robust cost function, the dense 3D reconstruction has been gradually improved as indicated by the decreasing 3D reconstruction errors.}
       \label{illustration}
\end{figure*}

In this paper, we look for a simple and elegant convex optimization for dense NRSfM that can be efficiently implemented on a CPU. We would like to argue that the inherent spatial and temporal smoothness constraints could be well exploited to regularize the dense non-rigid reconstruction problem. Specifically, we revisit the temporal smoothness in sparse reconstruction and demonstrate that it can be employed in dense case directly. Second, to exploit the spatial smoothness in dense reconstruction, we resort to the Laplacian of the 3D non-rigid shape, which captures the local smoothness and owns mathematical simplicity. Finally, to handle inevitable noise and outliers in real world image measurements, we robustify the data term by using the $L_1$ norm rather than commonly used $L_2$ norm. In this way, our method could robustly exploit the spatial-temporal smoothness in dense non-rigid reconstruction effectively. Our method is very easy to implement, which involves solving a series of least squares problems. In Fig.~\ref{illustration}, we demonstrate the contribution of each component. With the introduction of temporal smoothness, spatial smoothness and robust cost function, the dense 3D non-rigid reconstruction has been gradually improved.

% Our main contributions can be summarized as:
% \begin{enumerate}
% \item Revisit the temporal smoothness constraint and employ it to dense NRSfM;
% \item Exploit the spatial smoothness constraint by resorting to Laplacian matrix;
% \item Robustify the cost function with $L_1$ norm to deal with noise and outliers;
% \item Easy to implement solution with gradient descent.
% \end{enumerate}

%=============================================================
\section{Prerequisites}
Dense NRSfM takes a 2D video obtained by a monocular camera as input, with image frames $\m I_1 \cdots \m I_F$ each containing $P$ pixels. In this paper, we assume the per-pixel feature tracks have been extracted, say by optical flow or dense matching. Thus, the input to our system is a $2F\times P$ feature track matrix $\m W$, which stacks $F$ following $2\times P$ matrix:
$
\m W_i=\left[ 
\begin{array}{ccc}
u_{i1}, \cdots, u_{iP} \\
v_{i1}, \cdots, v_{iP} \\
\end{array}\right],
$
where $(u_{ij},v_{ij})^T$ denotes the $j$-th feature point captured in the $i$-th image frame. Assuming an orthographic camera model and the camera has been centralized at the center of the object, we have:
$
\m W_i=\m R_i\m S_i,
$
where $\m R_i$ is a $2\times 3$ matrix that represents the first two rows of the rotation matrix of the $i$-th frame, and $\m S_i$ is a $3\times P$ matrix containing the 3D positions of every point in the $i$-th non-rigid shape. Stacking all the feature tracks for all frames gives:
\vspace{0mm}
\begin{equation}
\m W=\m R \m S,
\end{equation}
\vspace{0mm}
with $\m R$ and $\m S$ are with dimension $2F\times 3F$ and $3F\times P$, respectively. Dense NRSfM aims at simultaneously recovering both the camera motion $\m R$ and the non-rigid shape $\m S$ from the feature track matrix $\m W$. The problem is inherently under-determined as the number of variables to estimate ($3FP+3F$) greatly exceeds the number of measurements $2FP$. Therefore, extra constraints are needed to regularize the problem. 

Under dense NRSfM, we solve for the camera rotation $\m R$ by utilizing the low-rank structure of $\m S$. Even though we have to deal with tens of thousands of points, the rotation estimation method in \cite{Dai-Li-He:IJCV-2013} still could handle it as the computational complexity is independent of the number of points but only depends on the model complexity $K$.

\section{Formulation and Solution}
In this paper, we propose to exploit the generic and generally available smoothness from temporal direction and spatial direction. By jointly enforcing the spatial and temporal constraints, we are able to achieve dense non-rigid reconstruction in an easy and elegant way.

\subsection{Temporal Smoothness Revisited}
First, we revisit the temporal smoothness, which has been widely used in sparse NRSfM \cite{Dai-Li-He:IJCV-2013}. We would like to argue that this simple strategy could be pretty efficient in achieving comparable performance with complex convex optimization or ADMM based methods.

By introducing smooth deformation regularization \cite{Fredrik02estimationof}\cite{Olsen-Bartoli:JMIV-2008} \cite{Lucey:CVPR-2012}, we can formulate the non-rigid shape recovery problem as minimizing a data term evaluated on the image measurements and a regularization term based on temporal smoothness, thus reaching the following optimization:
\begin{equation}
\label{eq:smooth_constraint} \min_{\m S} \frac{1}{2} \|\m W - \m R\m S\|_{\mathrm{F}}^2 + \frac{1}{2}\lambda\|\m H \m S\|_{\mathrm{F}}^2,
\end{equation}
where the first term measures the reprojection error evaluated on image plane while the second term measures the temporal smoothness constraint. We could apply different smooth operators $\m H$ to characterize various kinds of smoothness in temporal direction, \eg ~first order smoothness as in Eq.-\eqref{eq:first_order_smoothness}, second order smoothness and \etc.
\begin{equation}
\m H_{ij} = \left\{ \begin{array}{rl}
 1, & j=i,i=1,\cdots,3(F-1), \\
 -1,& j=i+3,i=1,\cdots,3(F-1),\\
 0, & \mathrm{Otherwise}.
\end{array} \right.
\label{eq:first_order_smoothness}
\end{equation}

The resultant optimization problem in Eq.-\eqref{eq:smooth_constraint} admits an analytical (closed-form) solution,
\begin{equation}
\label{eq:closed_form_solution} \m S_{\mathrm{smooth}} = (\m R^T \m R + \lambda \m H^T\m H)^{\dag}\m R^T\m W.
\end{equation}

The rotation matrix $\m R$ is of row full rank $2F$ thus $\m R^T\m R$ is of rank $2F$ generally.  The smoothness matrix $\m H$ is rank deficient too, thus $\m H^T\m H$ is of rank $3F-3$ (for first order smoothness).  In general case, $\m R^T\m R + \lambda \m H^T \m H$ is a full rank matrix, thus invertible.

The 3D non-rigid shape generated by this solution depends on the choice of the trade-off parameter $\lambda$, which trades off between 2D reprojection error and temporal smoothness. When $\lambda$ approaches 0, the solution approaches $\m R^{\dag} \m W$, \ie ~the pseudo-inverse solution.  When $\lambda$ is large enough, the solution approaches a rigid shape, which minimizes the combination of $\|\m H \m S\|_{\mathrm{F}}^2$ and $\| \m W - \m R \m S \|_{\mathrm{F}}^2$.  When $\lambda$ approaches $+\infty$, the solution approaches a trivial solution $\m S = \m 0$ \cite{Dai-Li-He:IJCV-2013}.

\textbf{Connection:} The smoothness constrained solution and the pseudo-inverse solution are connected as: 
\begin{equation}
\m S_{\mathrm{Smooth}} = (\m R^T \m R + \lambda \m H^T\m H)^{\dag}\m R^T\m W = (\m R^T \m R + \lambda \m H^T\m H)^{\dag} \m S_{\mathrm{PI}}.
\end{equation}
Therefore $\rank(\m S_{\mathrm{Smooth}}) = \rank(\m S_{\mathrm{PI}})$. As proved in \cite{Dai-Li-He:IJCV-2013}, the pseudo-inverse solution is a degenerate case where the non-rigid shape at each frame lies on a plane. $\m S_{Smooth}$ can be viewed as a per-frame weighted version of $\m S_{\mathrm{PI}}$.

%=============================================================
\subsection{Spatial Smoothness Simplified } %: Laplacian to the rescue
The temporal smoothness constrains the dense non-rigid reconstruction from the temporal dimension, \ie, the smoothness of 3D trajectory. However, it could not regularize the 3D shape at each frame. Garg \etal \cite{Dense-NRSFM:CVPR-2013} proposed to use the total variation to encourage the spatial smoothness while maintaining sharp boundaries. The resultant optimization prohibits its real world application to large scale 3D reconstruction.

\begin{figure}[!htp]
	\psfig{file=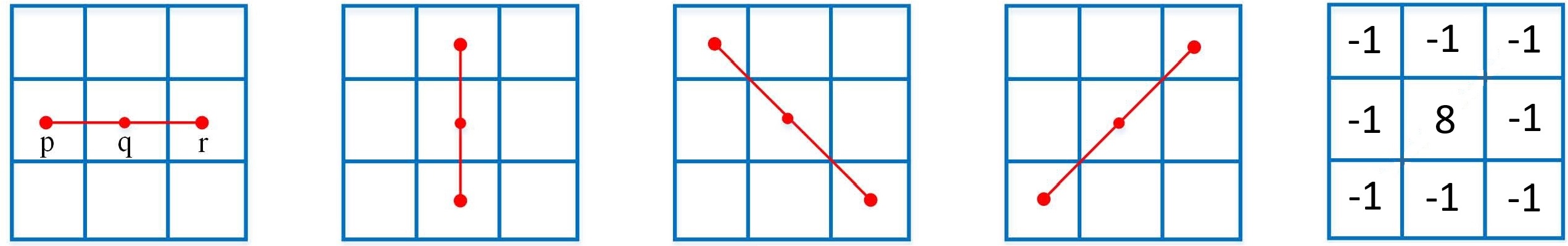,width=0.47\textwidth}
    \label{filter}
    \caption{Our Laplacian filter (far right): 8-direction, sum of the 4 basic Laplacian filters.}
\end{figure}

To efficiently and effectively utilize the smoothness alongside the spatial dimension, we propose a simple filtering mechanism, namely Laplacian filter, which enforces spatial smoothness locally in the 3D shape space. In Fig.~\ref{filter}, we illustrate different 2D filters in enforcing spatial smoothness. The Laplacian filter enforces a locally linear/planar model, which provides an easy way to encourage second order smoothness. As all linear filtering can be equivalently expressed as matrix multiplication, for the recovered non-rigid shape $\m S$, the filtering output is defined as:
\begin{equation}
\m A \mathrm{vec}(\m S),
\end{equation}
where $\m A$ is a $3FP \times 3FP$ matrix containing all the filtering operation, each row of $\m A$ defines a spatial filter evaluated at the position of $\m S_{ij}$.

Spatial smoothness is effective in smoothing 3D reconstruction. However, the spatial smoothness itself is not sufficient to recover the correct shape. Without temporal constraint, the result will be close to the pseudo-inverse case, which lies in a plane. By putting spatial smoothness and temporal smoothness together, we are able to achieve reliable 3D reconstruction even from noisy 2D inputs. 
% Here we present the performance of spatial-temporal smoothness on noisy data in Fig.~\ref{illustration}. Gaussian noise is added onto the 2D track with the standard deviation of $0.02\max\{|\m W|\}$. It can be observed that temporal smoothness alone cannot handle noise well, while spatial constraint combined with temporal constraint performs well in handling noisy data.

% \begin{figure}[!htp]
%     \subfigure{
%  \psfig{file=Final/Tnoise_syn2_L2_1.jpg,width=0.22\textwidth,height=0.16\textheight}
% 	   }
%     \subfigure{
%  \psfig{file=Final/STnoise_syn2_L2_1.jpg,width=0.22\textwidth,height=0.16\textheight}
% 	   }  
%     \subfigure{
%  \psfig{file=Final/Tnoise_syn2_L2_2.jpg,width=0.22\textwidth}
% 	   }
%     \subfigure{
%  \psfig{file=Final/STnoise_syn2_L2_2.jpg,width=0.22\textwidth}
% 	   }        
% \caption{Experimental results on noisy data. Left: temporal constraint only; Right: spatial-temporal constraint. Top: front view. Bottom: side view.}
% \label{ST_noise}
% \end{figure}

%============================================================
\subsection{Optimization Robustified } %: Corrupted Data
Noise and outliers are inevitable in real world measurements. Dense NRSfM methods must handle them robustly. Most of the existing methods apply $L_2$ on the data term, thus could not handle noise and outliers well. We propose to replace the $L_2$ norm with $L_1$ norm, thus increasing the robustness of the data term $\|\m W - \m R \m S \|_1 $.

To deal with the convex $L_1$ norm efficiently, we propose to use iterative reweighted least square (IRLS), where we solve for a least square problem in each iteration. Figure~\ref{illustration} illustrates the performance of $L_1$-norm on data with outliers. It is shown that our L1-norm relaxation gives a better performance in data with outliers.

% \begin{equation}
% \|\m W - \m R \m S \|_1 = \|\m E(\m W - \m R \m S)\|_F^2
% \end{equation}
% where $\m E$ is a $2F \times 2F$ diagonal matrix with $\m E_i=\frac{1}{\sqrt{\|\m W_i - \m R_i \m S_i^t\|}}$, where $\m S_i^t$ is the 3D shape obtained by the previous iteration. The above expression ensures that our optimization is still convex and has a closed-form solution. 

%=============================================================
\subsection{Spatial-Temporal smoothness constraint}
By enforcing the spatial-temporal smoothness constraint and applying the robust $L_1$ norm on data term, we reach:
\begin{equation}
\min_{\m S} \|\m W - \m R \m S \|_1 + \lambda_1 \|\m H \m S \|_F^2 + \lambda_2 \|\m A \mathrm{vec}(\m S) \|_F^2,
\end{equation}
where $\lambda_1$ and $\lambda_2$ are the trade-off parameters. The three terms are ``data term'', ``temporal smoothness term'' and ``spatial smoothness term'' correspondingly. Under IRLS formulation, we solve the following least square problem in each iteration:
\begin{equation}
\min_{\m S^{it}} \|\m E(\m W - \m R \m S) \|_F^2 + \lambda_1 \|\m H \m S \|_F^2 + \lambda_2 \|\m A \mathrm{vec}(\m S) \|_F^2.
\end{equation}
A closed-form solution can be derived by using the first order condition. However the computational complexity is high due to the filtering matrix $\m A$. Instead, we propose to solve the least square problem with gradient descent, where the gradient is derived as:
\begin{equation}
g(\m S) = 2 \m R^T \m E^T\m E\m R \m S - 2 \m R^T \m W + 2\lambda_1  \m H^T \m H \m S + 2\lambda_2 \mathrm{ivec} ((\m A^T \m A) \mathrm{vec}(\m S)),
\end{equation} $\mathrm{ivec}$ denotes the inverse operator of vectorization, which transforms a vector to matrix with proper dimension.
% The gradient descent is denoted as follows:

% \begin{equation}
% \m S^{(t+1)} = \m S^{(t)} - \mu g(\m S).
% \end{equation}
% where $\mu$ is the step parameter that increases each iteration. Gradient descent stops when the cost function  reaches a minimum value.

%=============================================================
\section{Experimental Results}
\textbf{Setting up:} To evaluate our method against existing state-of-the-art dense NRSfM methods, we used the 4 dense synthetic sequences and 3 real videos from \cite{Dense-NRSFM:CVPR-2013}. Each sequence contains a 2D correspondence matrix and a quad mesh for neighborhood assignment. These sequences have over 20,000 trajectories forming dense surfaces, which makes the problem much more challenging than the sparse scenarios.

We first enforced the temporal smoothness constraint to obtain initialized 3D non-rigid reconstruction. Then our method runs iteratively to optimize the cost function with spatial-temporal constraints. The trading-off parameters are set as $\lambda_1=10^{-3}$, $\lambda_2=1$.

\begin{figure}[!htb]
	   \subfigure{
       \psfig{file=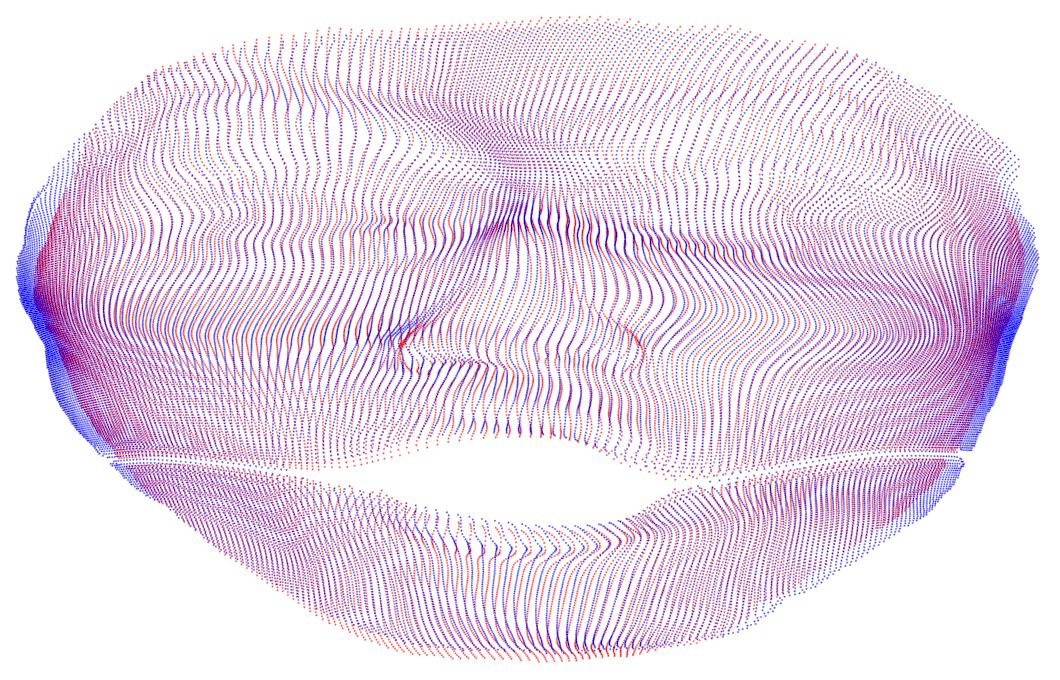,width=0.10\textwidth,
       height=0.08\textheight}
       \label{Syn1_ST_1}
	   }
       \subfigure{
       \psfig{file=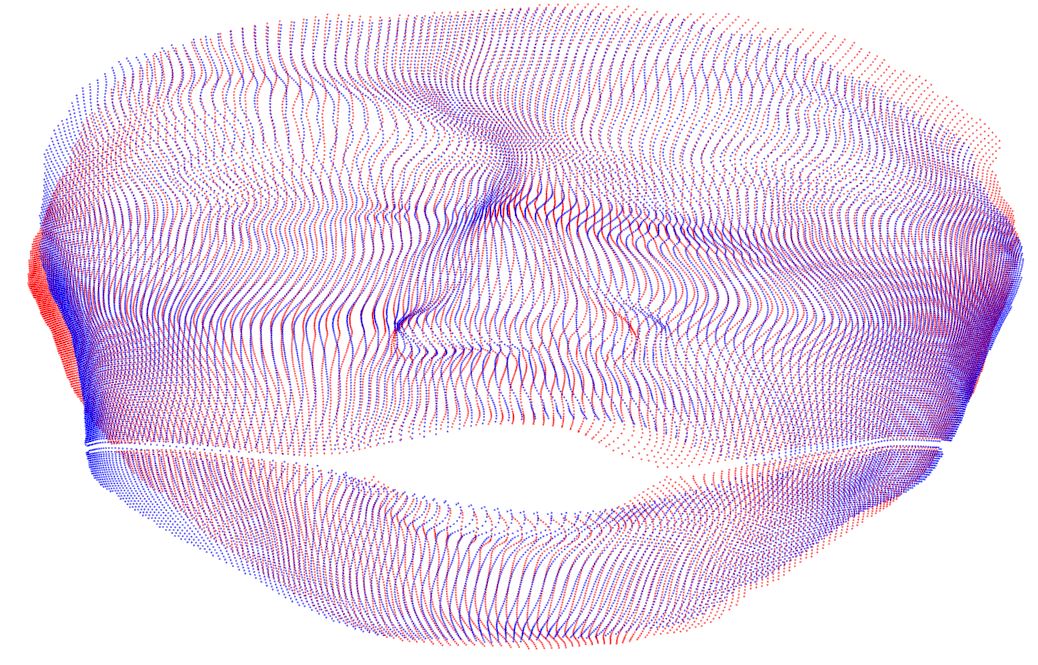,width=0.10\textwidth,
       height=0.08\textheight}
       \label{Syn2_ST_1}
	   }
       \subfigure{
       \psfig{file=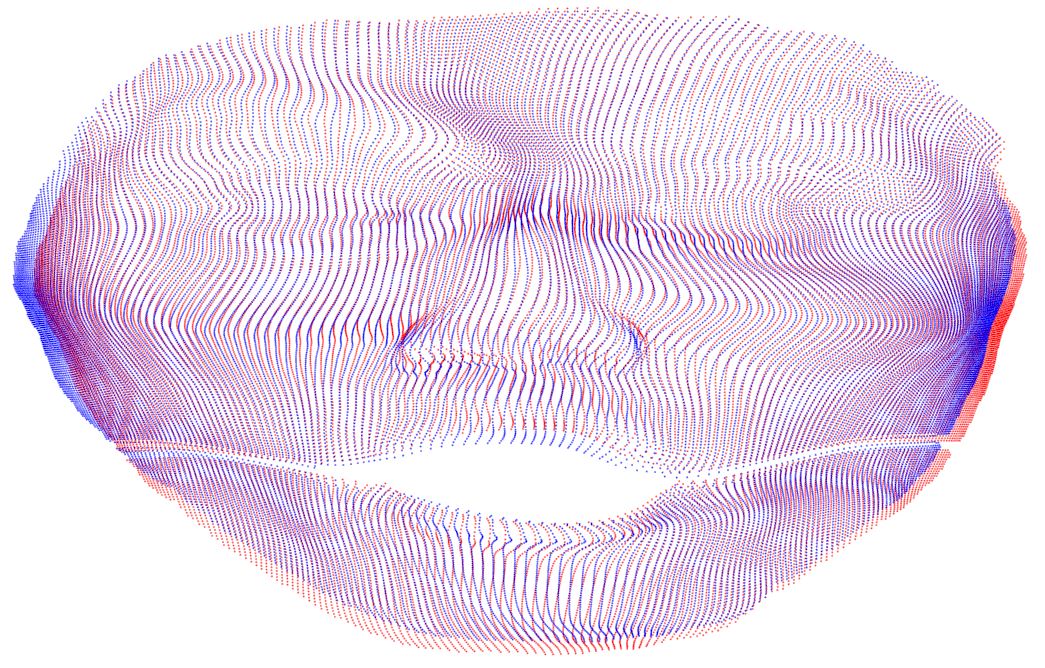,width=0.10\textwidth,
       height=0.08\textheight}
       \label{Syn3_ST_1}
	   }
       \subfigure{
       \psfig{file=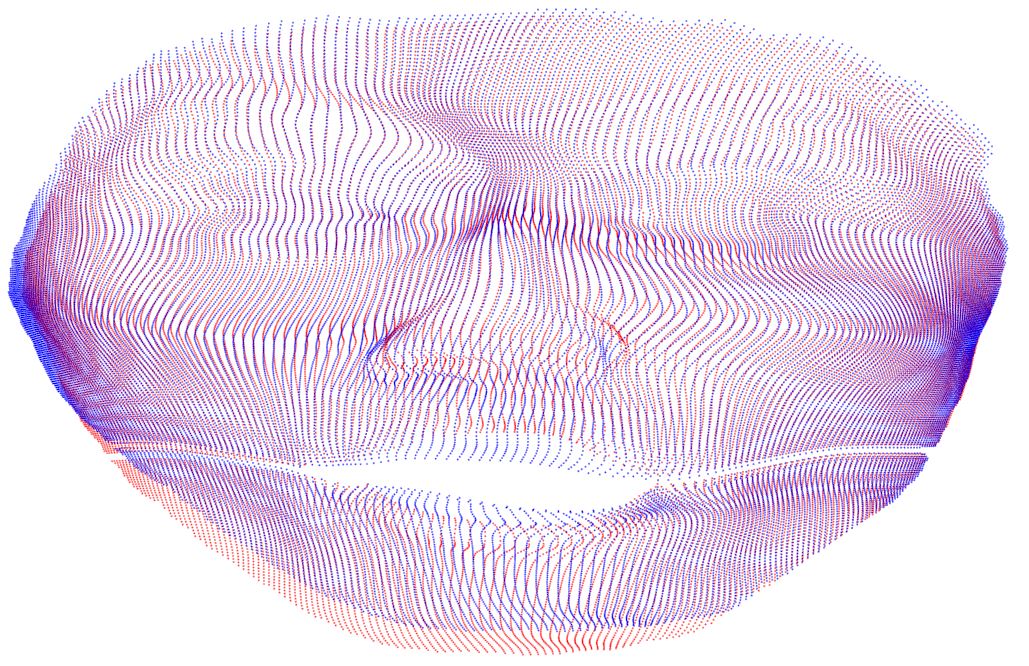,width=0.10\textwidth,
       height=0.08\textheight}
       \label{Syn4_ST_1}
	   }
       \setcounter{subfigure}{0}
       \subfigure[Seq 1]{
       \psfig{file=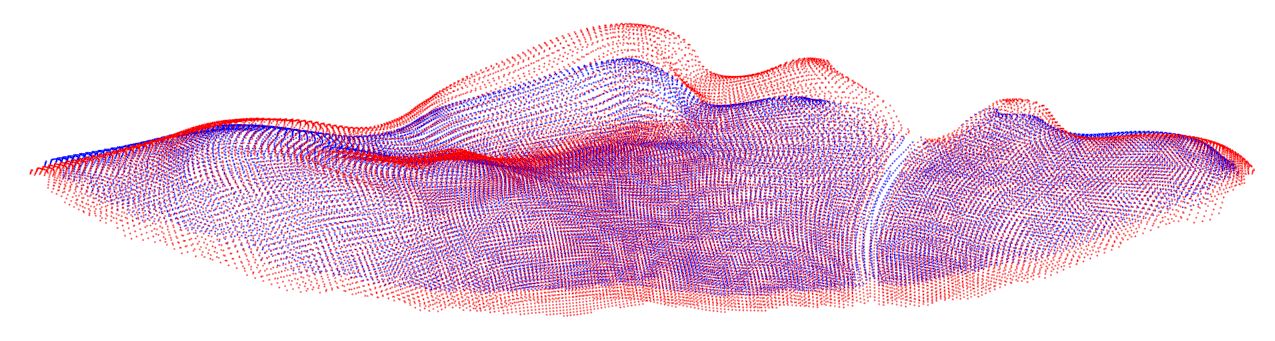,width=0.105\textwidth,
       height=0.025\textheight}
       \label{Syn1_ST_2}
	   }
       \subfigure[Seq 2]{
       \psfig{file=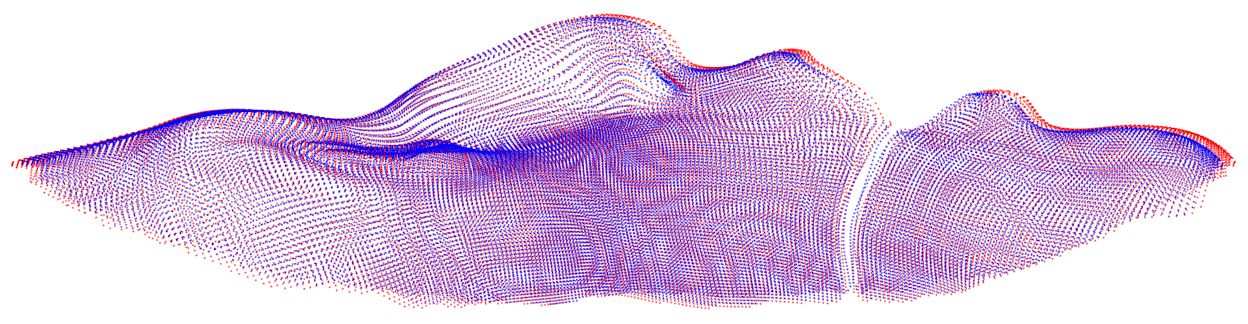,width=0.105\textwidth,
       height=0.025\textheight}
       \label{Syn2_ST_2}
	   }
       \subfigure[Seq 3]{
       \psfig{file=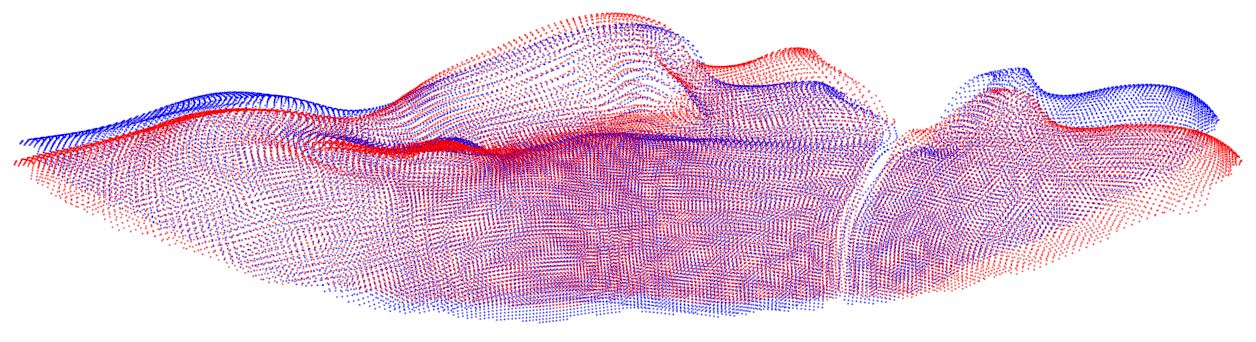,width=0.105\textwidth,
       height=0.025\textheight}
       \label{Syn3_ST_2}
	   }
       \subfigure[Seq 4]{
       \psfig{file=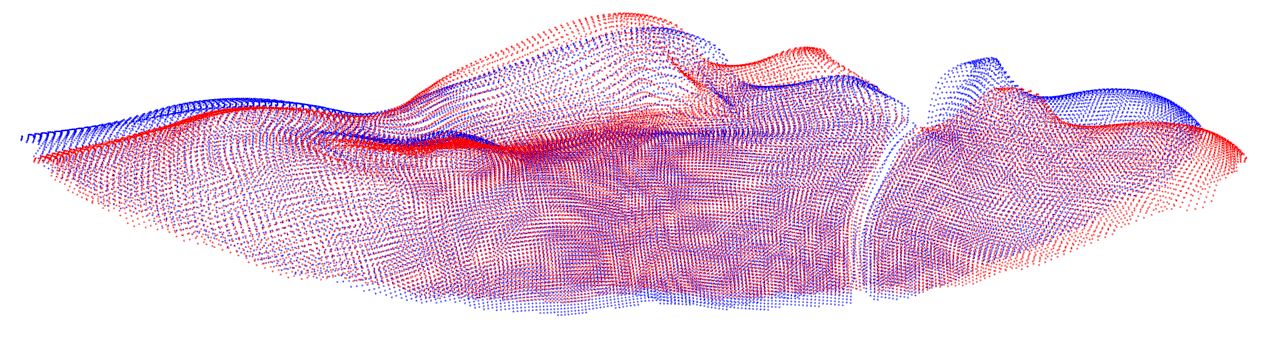,width=0.105\textwidth,
       height=0.025\textheight}
       \label{Syn4_ST_2}
	   }
       \caption{\small Dense non-rigid reconstruction results on synthetic face sequences. Red: ground truth; Blue: our results. Top row: front view. Bottom row: side view.}
       \label{Syn_ST}
\end{figure}

On synthetic face sequences, the results of our method are shown in Fig.~\ref{Syn_ST}. We overlap the ground truth shape in red and the our 3D reconstruction in blue. These figures show that our method can reconstruct the 3D object quite accurately. Table~\ref{tab:real_image_result_shape} shows the quantitative evaluation of our method along with various others methods, including Trajectory Basis (PTA)\cite{Akhter-Sheikh-Khan-Kanade:Trajectory-Space-2010}, Metric Projection\cite{metric-projection:CVPR-2009} and Variational method\cite{Dense-NRSFM:CVPR-2013}. As shown in the table, our method achieves competitive performance with the state-of-the-art methods. It is worth noting that our method is pretty easy to implement which only involves a series of least squares. 

%Especially for sequence 1, which is the most challenging, our method achieves good performance while both PTA and MP fail to reconstruct the object.

\begin{table}[!htp]
\caption{Quantitative evaluation on 4 synthetic face sequences. (Average RMS 3D reconstruction error.)}
 \centering
\begin{tabular}{|c| c| c|c| c| }\hline
 Dataset & PTA \cite{Akhter-Sheikh-Khan-Kanade:Trajectory-Space-2010}    & MP \cite{metric-projection:CVPR-2009}     & DV \cite{Dense-NRSFM:CVPR-2013} & Ours\\ \hline \hline
 Seq1    & 0.2431 & 0.2575  & 0.0531 & 0.0636\\  \hline
 Seq2    & 0.0988 & 0.0644  & 0.0457 & 0.0569 \\  \hline
 Seq3    & 0.0596 & 0.0682  & 0.0346 & 0.0374 \\  \hline
 Seq4    & 0.0877 & 0.0772  & 0.0379 & 0.0428 \\  \hline
\end{tabular}
\label{tab:real_image_result_shape}
\end{table}

\begin{figure}[!htb]
\centering
	\subfigure{
\psfig{file=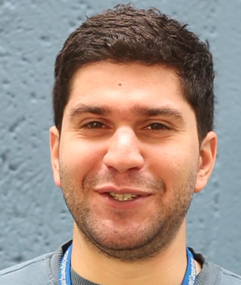,width=0.12\textwidth,height=0.11\textheight}
       \label{Face_ST_1}
	   }
       	\subfigure{
\psfig{file=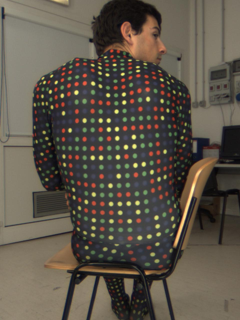,width=0.12\textwidth,height=0.11\textheight}
       \label{Face_ST_1}
	   }
       	\subfigure{
\psfig{file=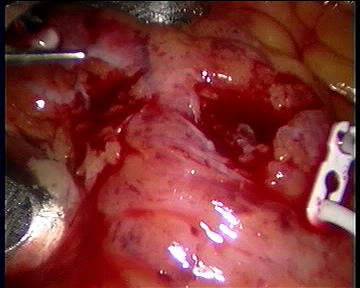,width=0.13\textwidth,height=0.11\textheight}
       \label{Face_ST_1}
	   }
      
	\subfigure{
\psfig{file=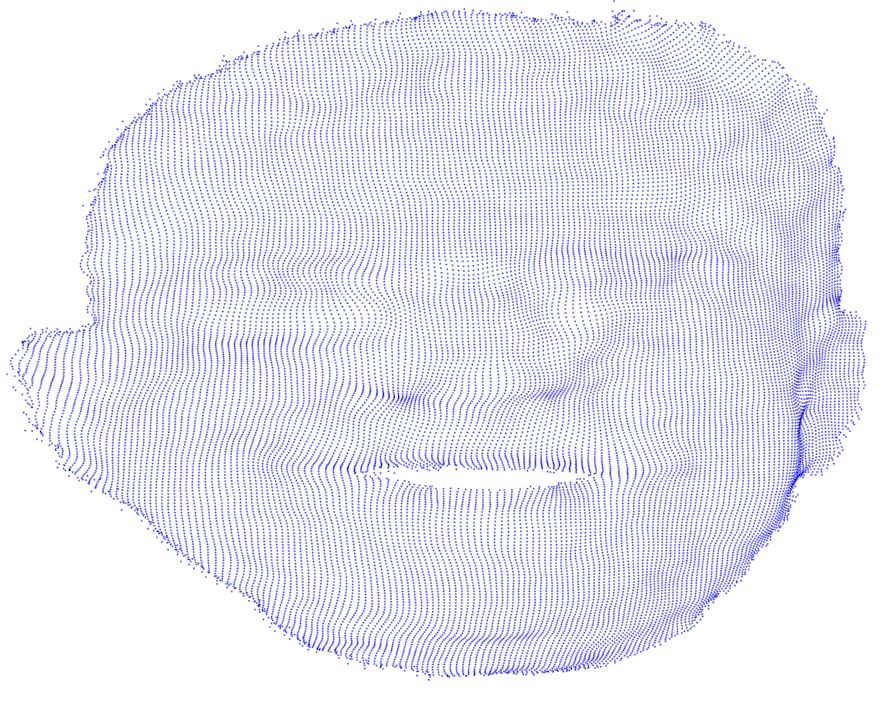,width=0.12\textwidth,height=0.1\textheight}
       \label{Face_ST_1}
	   }
       \subfigure{
       \psfig{file=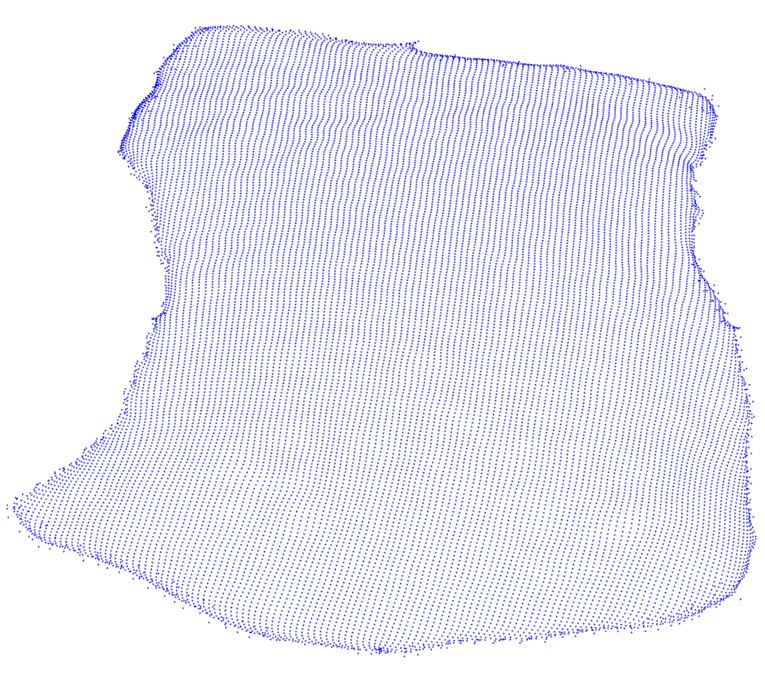,width=0.12\textwidth,height=0.1\textheight}
       \label{Back_ST_1}
	   }
       \subfigure{
       \psfig{file=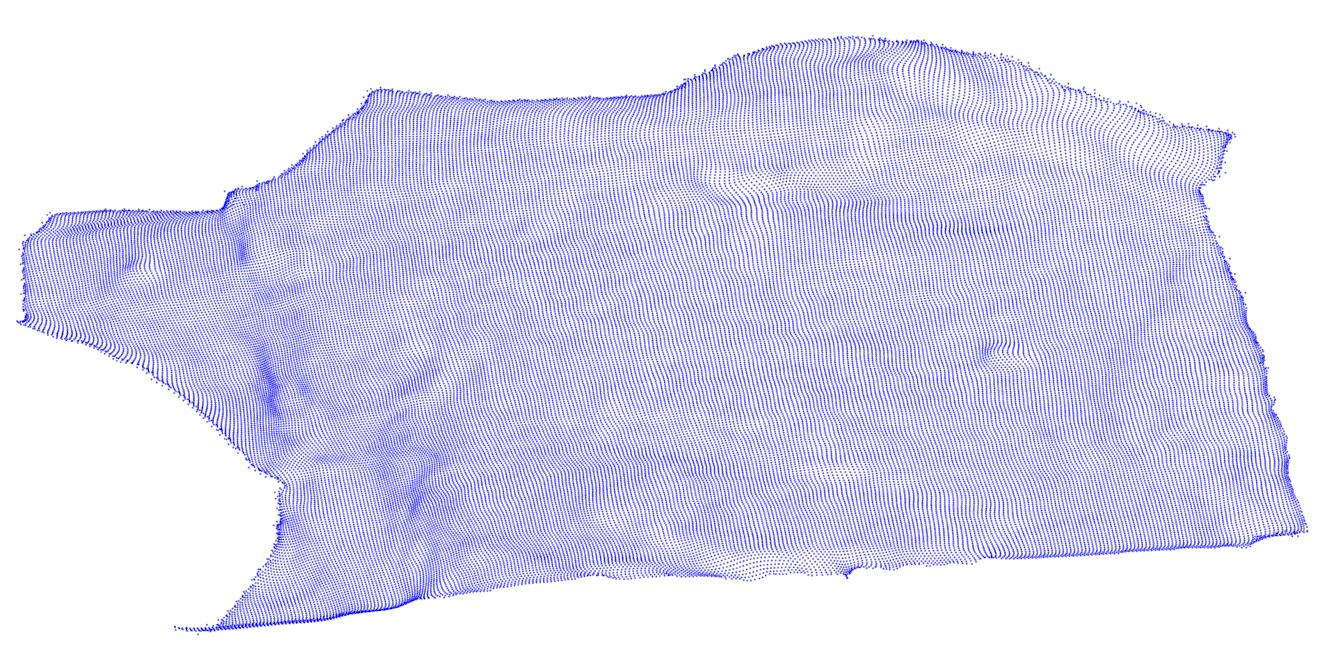,width=0.15\textwidth,height=0.07\textheight}       
       \label{Heart_T_1}
	   }
       \setcounter{subfigure}{0}
	   \subfigure[Face]{      \psfig{file=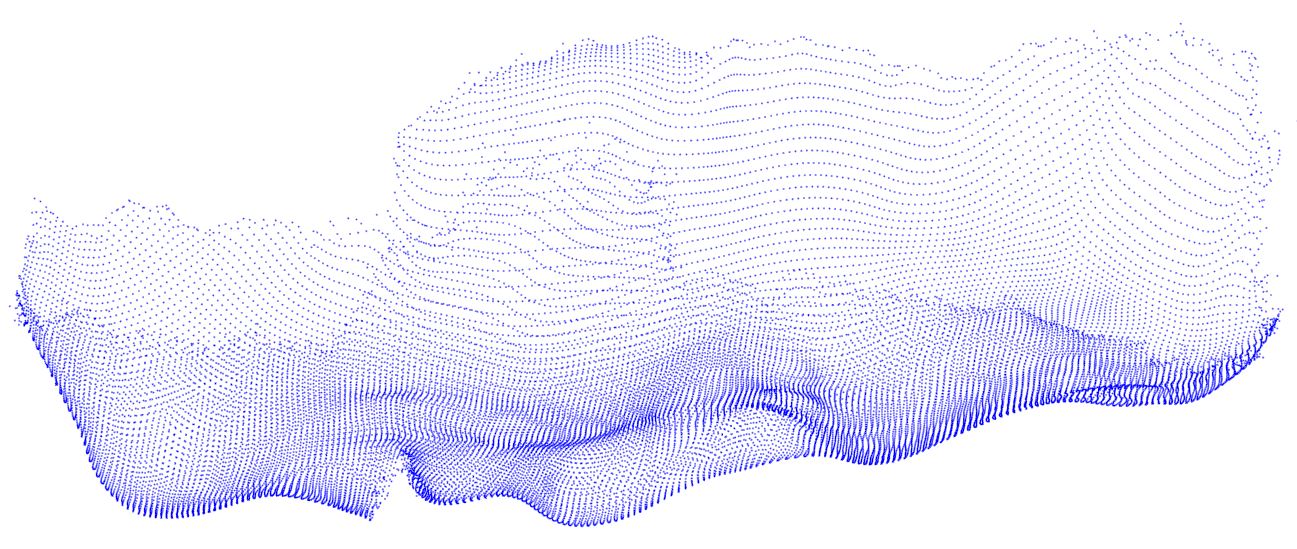,width=0.14\textwidth}
       \label{Face_ST_2}
	   }
       \subfigure[Back]{
       \psfig{file=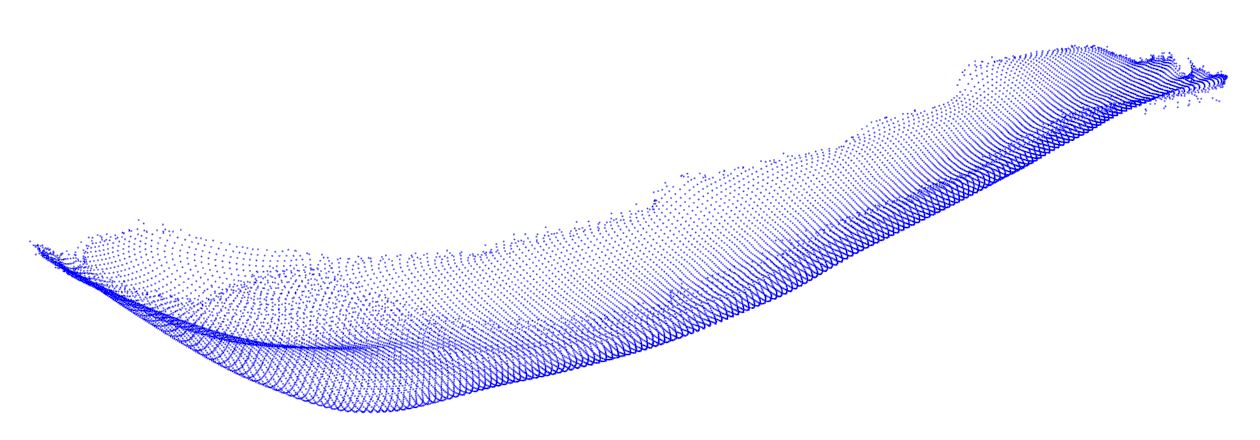,width=0.14\textwidth}
       \label{Back_ST_2}
	   }
       \subfigure[Heart]{
       \psfig{file=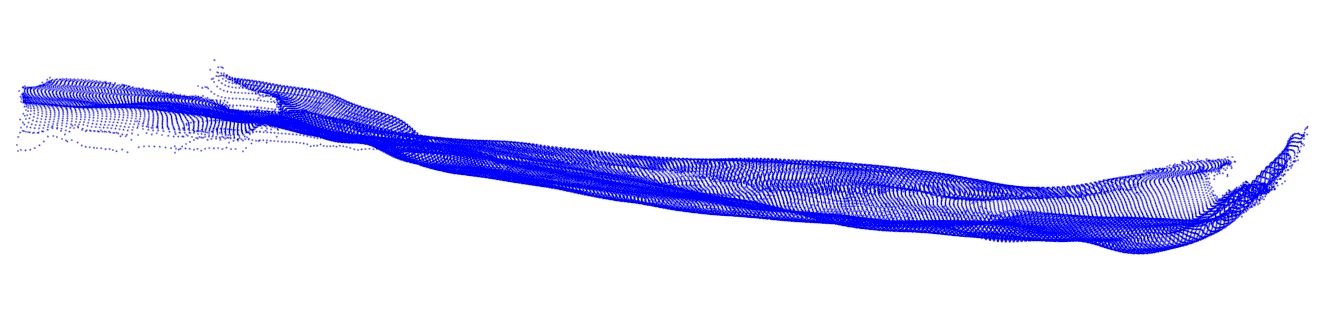,width=0.15\textwidth}
       \label{Heart_ST_2}
	   }
       \caption{\small Dense non-rigid 3D reconstruction results on the real image sequences (Face, Back and Heart). Top row: input 2D images. Middle row: front views of the respective sequences. Bottom row: side views.}
       \label{Real_ST}
\end{figure}

For dense sequences obtained from real videos, the input 2D video tracks and results obtained by our method are shown in Fig.~\ref{Real_ST}. As shown in the figures, our method outputs reasonable results on the Face and Back sequences, while on the challenging Heart sequence that has both large deformations and small rotation, our result seems to be too flat. This emphasizes the importance of a correct rotation matrix.

% \begin{figure*}[htb]
% 	   \subfigure{
%        \psfig{file=Final/Face_Wr.JPG,height=0.085\textheight}
%        \label{Face_W}
% 	   }
%        \subfigure{
%        \psfig{file=Final/Back_Wr.JPG,height=0.085\textheight}
%        \label{Back_W}
% 	   }
%        \subfigure{
%        \psfig{file=Final/Heart_Wr.JPG,height=0.085\textheight}
%        \label{Heart_W}
% 	   }
%        \caption{\small Real 2D videos from left to right: Face, Back and Heart, respectively.}
%        \label{Real_W}
% \end{figure*}

\textbf{Dense input with noise:}
As stated in previous sections, assuming a smooth 3D surface, our spatial smoothness constraint encourages local smoothness, hence increasing the accuracy and resolution. To evaluate the performance of our method, we added Gaussian noise to the 2D input images, with the standard deviation $\sigma_n=r\max\{|\m W|\}$, where $r$ is the noise ratio ranging from 0.01 to 0.05. Each noise settings are repeated for 5 times to obtain statistical results.

\begin{figure}[!htb]
	   \subfigure[]{    \psfig{file=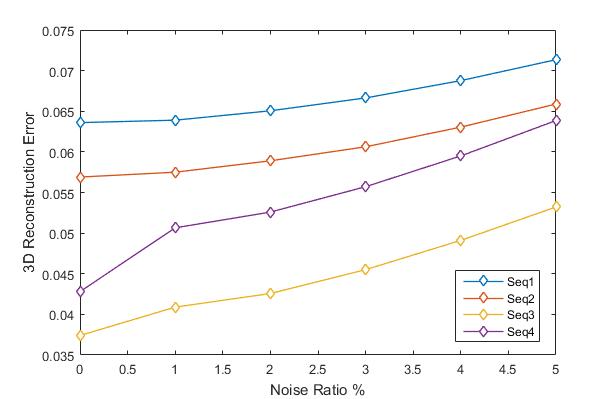,width=0.225\textwidth}
       \label{noise}
       }
       \subfigure[]{    \psfig{file=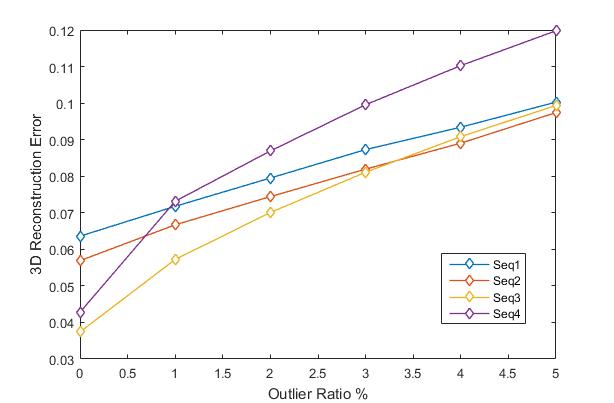,width=0.225\textwidth}
       \label{outlier}
	   }
       \caption{Performance evaluation under noise and outliers. (a) Experimental results (3D error) w.r.t noise levels. (b) 3D reconstruction error w.r.t outlier ratios.}     
\end{figure}

Figure ~\ref{noise} shows the performance of our method under different noise ratios on 4 synthetic sequences. It shows that even at large noise ratios, the 3D error of our method is still kept at a low level. 

\textbf{Dense input with outliers:} To evaluate the capability of our method in dealing with outliers, we performed experiments with the following settings: a certain amount of points in the video ($FP$ points in total) are set at random positions. The outlier ratio varies at 2\%, 4\%, 6\%, 8\% and 10\%, respectively. We compute the final 3D error by averaging 5 trials, in order to get a statistically accurate result.

Figure ~\ref{outlier} illustrates the performance of our method under different outlier ratios. As outlier ratio increases, the 3D error increases slightly, keeping under 0.1 for all synthetic sequences. The error curves are almost linear, which demonstrates the robustness of our method.

%=============================================================
\section{Conclusions}
In this paper, we propose a unified framework to dense non-rigid 3D reconstruction, which utilizes both spatial and temporal smoothness to regularize the under-constrained problem. Furthermore, the cost function has been robustified to deal with real world noise and outliers. Our method achieves competitive performance with state-of-the-art dense NRSfM methods. The implementation of our method only involves solving a series of least squares problems, thus making dense NRSfM easy. 

%In future, we plan to investigate the theoretical aspect of dense NRSfM, \ie, the reconstructionability, and performance bound. In this paper, we solve for the camera rotation first. We also plan to include the  camera rotation in the optimization, thus further improving the result.

\bibliographystyle{IEEEbib}
\bibliography{Reference}

\end{document}